\documentclass[11pt,a4paper]{article}
\usepackage[hyperref]{naaclhlt2018}
\usepackage{times}
\usepackage{latexsym}

\usepackage{pgfplots}
\usepackage{tikz-qtree}
\usepackage{tikz-dependency}
\usepackage{url}
\usepackage{graphicx}
\graphicspath{{images/}}
\usepackage{amssymb,amsthm, amsmath}

\renewcommand{\underbar}[1]{#1}

\aclfinalcopy 

\title{End-to-end Graph-based TAG Parsing with Neural Networks}

\author{Jungo Kasai\textsuperscript{$\clubsuit$} \quad\quad\quad Robert Frank\textsuperscript{$\clubsuit$} \quad\quad\quad Pauli Xu\textsuperscript{$\clubsuit$} \quad \quad \quad \\
{\bf William Merrill\textsuperscript{$\clubsuit$}} \quad\quad\quad {\bf Owen Rambow\textsuperscript{$\heartsuit$}} \quad\quad\quad\\
\textsuperscript{$\clubsuit$}Department of Linguistics, Yale University\\
\textsuperscript{$\heartsuit$}Elemental Cognition, LLP\\
\scalebox{0.85}[0.9]{{\tt \{jungo.kasai,bob.frank,pauli.xu,william.merrill\}@yale.edu}}\\
\scalebox{0.85}[0.9]{{\tt owenr@elementalcognition.com}}}

\date{}

\begin{document}
\maketitle
\begin{abstract}
We present a graph-based Tree Adjoining Grammar (TAG) parser that uses BiLSTMs, highway connections, and character-level CNNs.
Our best end-to-end parser, which jointly performs supertagging, POS tagging, and parsing, outperforms the previously reported best results by more than 2.2 LAS and UAS points.
The graph-based parsing architecture allows for global inference and rich feature representations for TAG parsing, alleviating the fundamental trade-off between transition-based and graph-based parsing systems.
We also demonstrate that the proposed parser achieves state-of-the-art performance in the downstream tasks of Parsing Evaluation using Textual Entailments (PETE) and Unbounded Dependency Recovery. This provides further support for the claim that TAG is a viable formalism for problems that require rich structural analysis of sentences.
\end{abstract}

\section{Introduction}
Tree Adjoining Grammar (TAG, \citet{JoshiSchabes97}) and Combinatory Categorial Grammar (CCG, \citet{SteedmanBaldridge11}) are both mildly context-sensitive grammar formalisms that are lexicalized: every elementary structure (elementary tree for TAG and category for CCG) is associated with exactly one lexical item, and every lexical item of the language is associated with a finite set of elementary structures in the grammar \cite{rambow/joshi:1994b}.
In TAG and CCG, the task of parsing can be decomposed into two phases (e.g. TAG: \citet{Bangalore1999}; CCG: \citet{clark2007wide}): \textit{supertagging}, where elementary units or \textit{supertags} are assigned to each lexical item and \textit{parsing} where these supertags are combined together.
The first phase of supertagging can be considered as ``almost parsing" because supertags for a sentence almost always determine a unique parse \cite{Bangalore1999}.
This near uniqueness of a parse given a gold sequence of supertags has been confirmed empirically (TAG: \citet{bangalore-EtAl:2009:NAACLHLT09-Short, chung:etal:TAG+12:2016, kasai-EtAl:2017:EMNLP2017}; CCG: \citet{lewis2016lstm}).

We focus on TAG parsing in this work. 
TAG differs from CCG in having a more varied set of supertags.
Concretely, the TAG-annotated version of
the WSJ Penn Treebank \cite{Marcus:1993:BLA:972470.972475} that we use \cite{chen05} includes
4727 distinct supertags (2165 occur once)
while the CCG-annotated version \cite{HockenmaierSteedman07} only includes 1286 distinct supertags
(439 occur once). 
This large set of supertags in TAG presents a severe challenge in supertagging and causes a large discrepancy in parsing performance with gold supertags and predicted supertags \cite{bangalore-EtAl:2009:NAACLHLT09-Short, chung:etal:TAG+12:2016, kasai-EtAl:2017:EMNLP2017}.

In this work, we present a supertagger and a parser that substantially improve upon previously reported results.
We propose crucial modifications to the bidirectional LSTM (BiLSTM) supertagger in \citet{kasai-EtAl:2017:EMNLP2017}. First, we use character-level Convolutional Neural Networks (CNNs) for encoding morphological information instead of suffix embeddings. 
Secondly, we perform concatenation after each BiLSTM layer.
Lastly, we explore the impact of adding additional BiLSTM layers and highway connections.
These techniques yield an increase of 1.3\% in accuracy. 
For parsing, since the derivation tree in a lexicalized TAG is a type of dependency tree \cite{rambow/joshi:1994b}, we can directly apply dependency parsing models. In particular, we use the biaffine graph-based parser proposed by \citet{dozatmanning2017} together with
our novel techniques for supertagging.

In addition to these architectural extensions for supertagging and parsing, we also explore multi-task learning approaches for TAG parsing.
Specifically, we perform POS tagging, supertagging, and parsing using the same feature representations from the BiLSTMs. This joint modeling has the benefit of avoiding a time-consuming and complicated pipeline process, and instead produces a full syntactic analysis, consisting of supertags and the derivation that combines them, simultaneously.
Moreover, this multi-task learning framework further improves performance in all three tasks. 
We hypothesize that our multi-task learning yields feature representations in the LSTM layers that are more linguistically relevant  and that generalize better \cite{caruana1997}. 
We provide support for this hypothesis by analyzing syntactic analogies across induced vector representations of supertags \cite{kasai-EtAl:2017:EMNLP2017, friedman-EtAl:2017:TAG+13}.
The end-to-end TAG parser substantially outperforms the previously reported best results.

Finally, we apply our new parsers to the downstream tasks of Parsing Evaluation using Textual Entailements (PETE, \citet{yuret-han-turgut:2010:SemEval}) and Unbounded Dependency Recovery \cite{unbounded-dependencies}. We demonstrate that our end-to-end parser outperforms the best results 
in both tasks.
These results illustrate that TAG is a viable formalism for  tasks that benefit from the assignment of rich structural descriptions to sentences.

\section{Our Models}
TAG parsing can be decomposed into \textit{supertagging} and \textit{parsing}.
Supertagging assigns to words elementary trees (supertags) chosen from a finite set, and parsing determines how these elementary trees can be combined to form a derivation tree that yield the observed sentence.
The combinatory operations consist of {\em substitution}, which  inserts obligatory arguments, and {\em adjunction}, which is responsible for the introduction of modifiers, function words, as well as the derivation of sentences involving long-distance dependencies.
In this section, we present our supertagging models, parsing models, and joint models.
\subsection{Supertagging Model}
Recent work has explored neural network models for supertagging in TAG \cite{kasai-EtAl:2017:EMNLP2017} and CCG \cite{xu-auli-clark:2015:ACL-IJCNLP, lewis2016lstm, vaswani-EtAl:2016:N16-1, xu:2016:EMNLP2016}, and has shown that such models  substantially improve performance beyond non-neural models. 
We extend previously proposed BiLSTM-based models  \cite{lewis2016lstm,kasai-EtAl:2017:EMNLP2017} in three ways: 1) we add character-level Convolutional Neural Networks (CNNs) to the input layer, 2) we perform concatenation of both directions of the LSTM not only after the final layer but also after each layer, and 3) we use a modified BiLSTM with highway connections.
\subsubsection{Input Representations}
The input for each word is represented via concatenation of  a 100-dimensional embedding of the word, a 100-dimensional embedding of a predicted part of speech (POS) tag, and a 30-dimensional character-level representation from CNNs that have been found to  capture morphological information \cite{icml2014c2_santos14,TACL792,ma-hovy:2016:P16-1}. 
The CNNs encode each character in a word by a 30 dimensional vector and 30 filters produce a 30 dimensional vector for the word.
We initialize the word embeddings to be the pre-trained GloVe vectors \cite{pennington2014glove}; for words not in GloVe,
we initialize their embedding to a zero vector. 
The other embeddings are randomly initialized.
We obtain predicted POS tags from a BiLSTM POS tagger with the same configuration as in \newcite{ma-hovy:2016:P16-1}. 

\subsubsection{Deep Highway BiLSTM}
\label{highwaybilstm}
The core of the supertagging  model is a deep bidirectional Long Short-Term Memory network \cite{graves2005bilstm}. We use the following formulas to compute the activation of a single LSTM cell at time step $t$:
\begin{align}
    i_t &= \sigma \left (W_i [x_t ; h_{t-1}] + b_i \right )\\
    f_t &= \sigma \left (W_f [x_t ; h_{t-1}] + b_f \right )\\
    \tilde{c}_t &= \tanh \left (W_c [x_t ; h_{t-1}] + b_c \right )\\
    o_t &= \sigma \left (W_o [x_t ; h_{t-1}] + b_o \right )\\
    c_t &= f \odot c_{t-1} + i_t \odot \tilde{c}_t\\
    h_t &= o \odot \tanh \left ( c_t \right )
    \label{lstm}
\end{align}
Here a semicolon $;$ means concatenation, $\odot$ is element-wise multiplication, and $\sigma$ is the sigmoid function. In the first BiLSTM layer, the input $x_t$ is the vector representation of word $t$. (The sequence is reversed for the backwards pass.) In all subsequent layers, $x_t$ is the corresponding output from the previous BiLSTM; the output of a BiLSTM at timestep $t$ is equal to $[h_{t}^f ; h_{t}^b]$, the concatenation of hidden state corresponding to input $t$ in the forward and backward pass. 
This concatenation after each layer differs from \citet{kasai-EtAl:2017:EMNLP2017} and \citet{lewis2016lstm}, where concatenation happens only after the final BiLSTM layer.
We will show in a later section that concatenation after each layer contributes to improvement in performance.

We also extend the models in \citet{kasai-EtAl:2017:EMNLP2017} and \citet{lewis2016lstm} by allowing highway connections between LSTM layers. A highway connection is a gating mechanism that combines the current and previous layer outputs, which can prevent the problem of vanishing \!/\! exploding gradients \cite{srivastava2015highway}. Specifically, in networks with highway connections, we replace Eq.\ \ref{lstm} by:
\begin{align*}
    r_t &= \sigma \left (W_r [x_t ; h_{t-1}] + b_r \right )\\
    h_t &= r_t \odot o_t \odot \tanh \left ( c_t \right ) + \left (1 - r_t \right) \odot W_h x_t
\end{align*}
Indeed, our experiments will show that highway connections play a crucial role as we add more BiLSTM layers.

We generally follow the hyperparameters chosen in \newcite{lewis2016lstm} and \newcite{kasai-EtAl:2017:EMNLP2017}.
Specifically, we use BiLSTMs layers with 512 units each. Input, layer-to-layer, and recurrent \cite{Gal2016Theoretically} dropout rates are all 0.5.
For the CNN character-level representation, we used the hyperparameters from \newcite{ma-hovy:2016:P16-1}.

We train this network, including the embeddings, by optimizing the negative
log-likelihood of the observed sequences of
supertags in a mini-batch stochastic fashion with the Adam optimization algorithm with batch size 100 and $\ell = 0.01$ \cite{Kingma2015}.
In order to obtain predicted POS tags and supertags of the training data for subsequent parser input, we also perform 10-fold jackknife training.
After each training epoch, we test the supertagger on the dev set. 
When classification accuracy does not improve on five consecutive epochs, training ends.

\subsection{Parsing Model}
Until recently, TAG parsers have been grammar based, requiring as input a set of elemenetary trees (supertags).  For example, \citet{bangalore-EtAl:2009:NAACLHLT09-Short} proposes the MICA parser, an Earley parser that exploits a TAG grammar that has been transformed into a variant of a probabilistic CFG. One advantage of such a parser is that its parses are guaranteed to be well-formed according to the TAG grammar provided as input.

More recent work, however, has shown that data-driven transition-based parsing systems outperform such grammar-based parsers \cite{chung:etal:TAG+12:2016, kasai-EtAl:2017:EMNLP2017, friedman-EtAl:2017:TAG+13}. \citet{kasai-EtAl:2017:EMNLP2017} and \citet{friedman-EtAl:2017:TAG+13} achieved state-of-the-art TAG parsing performance using an unlexicalized shift-reduce parser with feed-forward neural networks that was trained on a version of the Penn Treebank that had been annotated with TAG derivations. 
Here, we pursue this data-driven approach, applying a graph-based parser with deep biaffine attention \cite{dozatmanning2017} that allows for global training and inference.


\subsubsection{Input Representations}
The input for each word is the concatenation of  a 100-dimensional embedding of the word and a 30-dimensional character-level representation obtained from CNNs in the same fashion as in the supertagger.\footnote{We fix the embedding of the ROOT token to be a 0-vector.}
We also consider adding 100-dimensional embeddings for a predicted POS tag \cite{dozatmanning2017} and a predicted supertag \cite{kasai-EtAl:2017:EMNLP2017, friedman-EtAl:2017:TAG+13}. 
The ablation experiments in \citet{KG2016} illustrated that adding predicted POS tags boosted performance in Stanford Dependencies.
In Universal Dependencies, \citet{dozat-qi-manning:2017:K17-3} empirically showed that their dependency parser gains significant improvements by using POS tags predicted by a Bi-LSTM POS tagger.
Indeed, \citet{kasai-EtAl:2017:EMNLP2017} and \citet{friedman-EtAl:2017:TAG+13} demonstrated that their unlexicalized neural network TAG parsers that only get as input predicted supertags can achieve state-of-the-art performance, with lexical inputs providing no improvement in performance.
We initialize word embeddings to be the pre-trained GloVe vectors as in the supertagger.
The other embeddings are randomly initialized.

\subsubsection{Biaffine Parser}
\begin{figure}
\hspace{-4mm}
    \centering
    \includegraphics[width=0.48\textwidth]{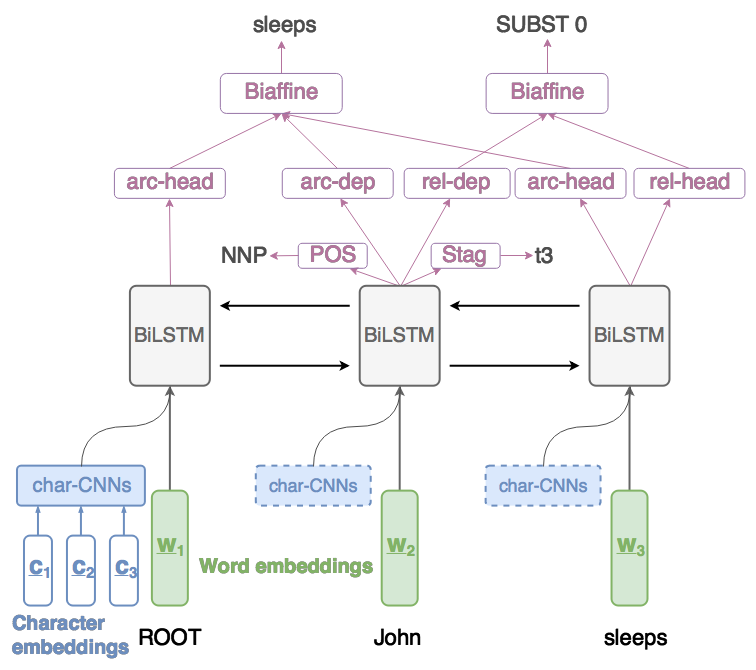}
    \caption{Biaffine parsing architecture. For the dependency from \textit{John} to \textit{sleeps} in the sentence \textit{John sleeps}, the parser first predicts the head of \textit{John} and then predicts the dependency label by combining the dependent and head representations. In the joint setting, the parser also predicts POS tags and supertags.}
    \label{biaffine}
\end{figure}
We train our parser to predict edges between lexical items in an LTAG derivation tree. 
Edges are labeled by the operations together with the deep syntactic roles of substitution sites (0=underlying subject, 1=underlying direct object, 2=underlying indirect object, 3,4=oblique arguments, CO=co-head for prepositional/particle verbs, and adj=all adjuncts).
Figure \ref{biaffine} shows our biaffine parsing architecture.
Following \citet{dozatmanning2017} and \citet{KG2016}, we use BiLSTMs to obtain features for each word in a sentence. 
We add highway connections in the same fashion as our supertagging model.

We first perform unlabeled arc-factored scoring using the final output vectors from the BiLSTMs, and then label the resulting arcs. 
Specifically, suppose that we score edges coming into the $i$th word in a sentence i.e. assigning scores to the potential parents of the $i$th word. 
Denote the final output vector from the BiLSTM for the $k$th word by $h_k$ and suppose that $h_k$ is $d$-dimensional.
Then, we produce two vectors from two separate multilayer perceptrons (MLPs) with the ReLU activation:
\begin{align*}
   &h_k^{\text{arc-dep}}  = \text{MLP}^{\text{(arc-dep)}} (h_k)\\
   &h_k^{\text{arc-head}}  = \text{MLP}^{\text{(arc-head)}} (h_k)
\end{align*}
where $h_k^{\text{arc-dep}}$ and $h_k^{\text{arc-head}}$ are $d_{arc}$-dimensional vectors that represent the $k$th word as a dependent and a head respectively.
Now, suppose the kth row of matrix $H^{\text{(arc-head)}}$ is $h_k^{\text{arc-head}}$. Then, the probability distribution $s_i$ over the potential heads of the $i$th word is computed by 
\begin{align}
\begin{split}
s_i 
= \text{softmax}(H^{\text{(arc-head)}} W^{\text{(arc)}}h_i^{\text{arc-dep}}\\
+H^{\text{(arc-head)}}b^{\text{(arc)}})
\label{unlabeled:eq}
\end{split}
\end{align}
where $W^{\text{(arc)}} \in \mathbb{R}^{d_{arc} \times d_{arc}}$ and $b^{(arc)} \in \mathbb{R}^{d_{arc}}$.
In training, we simply take the greedy maximum probability to predict the parent of each word.
In the testing phase, we use the heuristics formulated by \citet{dozatmanning2017} to ensure that the resulting parse is single-rooted and acyclic.

Given the head prediction of each word in the sentence, we assign labeling scores using  vectors obtained from two additional MLP with ReLU. For the $k$th word, we obtain:
\begin{align*}
   &h_k^{\text{rel-dep}}  =\text{MLP}^{\text{(rel-dep)}} (h_k)\\
   &h_k^{\text{rel-head}}  = \text{MLP}^{\text{(rel-head)}} (h_k)
\end{align*}
where $h_k^{\text{rel-dep}}, h_k^{\text{rel-head}} \in \mathbb{R}^{d_{rel}}$.
Let  $p_i$ be the index of the predicted head of the $i$th word, and  $r$ be the number of dependency relations in the dataset. 
Then, the probability distribution $\ell_i$ over the possible dependency relations of the arc pointing from the $p_i$th word to the $i$th word is calculated by:
\begin{align}
\begin{split}
   \ell_i = \text{softmax}(h_{p_i}^{T\text{(rel-head)}}U^{\text{(rel)} }h_{i}^{\text{(rel-dep)}}\\
   + W^{\text{(rel)}}(h_{i}^{\text{(rel-head)}}+h_{p_i}^{\text{(rel-head)}}) + b^{\text{(rel)}})
\label{labeled:eq}
\end{split}
\end{align}
where $U^{\text{(rel)}} \in \mathbb{R}^{d_{rel} \times d_{rel}\times r}$, $W^{\text{(rel)}} \in \mathbb{R}^{r \times d_{rel}}$, and $b^{\text{(rel)}} \in \mathbb{R}^{r}$.

We generally follow the hyperparameters chosen in \newcite{dozatmanning2017}.
Specifically, we use BiLSTMs layers with 400 units each. Input, layer-to-layer, and recurrent dropout rates are all 0.33.
The depths of all MLPs are all 1, and the MLPs for unlabeled attachment and those for labeling contain 500 ($d_{arc}$) and 100 ($d_{rel}$) units respectively. For character-level CNNs, we use the hyperparameters from \newcite{ma-hovy:2016:P16-1}.

We train this model with the Adam algorithm to minimize the sum of the cross-entropy losses from head predictions ($s_i$ from Eq.\ \ref{unlabeled:eq}) and label predictions ($\ell_i$ from Eq.\ \ref{labeled:eq}) with $\ell=0.01$ and batch size 100 \cite{Kingma2015}. 
After each training epoch, we test the parser on the dev set. 
When labeled attachment score (LAS)\footnote{We disregard pure punctuation when evaluating LAS and UAS, following  prior work \cite{bangalore-EtAl:2009:NAACLHLT09-Short, chung:etal:TAG+12:2016, kasai-EtAl:2017:EMNLP2017, friedman-EtAl:2017:TAG+13}.} does not improve on five consecutive epochs, training ends.

\subsection{Joint Modeling}
The simple BiLSTM feature representations for parsing presented above are conducive to joint modeling of POS tagging and supertagging; rather than using POS tags and supertags to predict a derivation tree, we can instead use the BiLSTM hidden vectors derived from lexical inputs alone to predict POS tags and supertags along with the TAG derivation tree.
\begin{align*}
   &h_k^{\text{pos}}  = \text{MLP}^{\text{(pos)}} (h_k)\\
   &h_k^{\text{stag}}  = \text{MLP}^{\text{(stag)}} (h_k)
\end{align*}
where $h_k^{\text{pos}}\in \mathbb{R}^{d_{pos}}$ and $h_k^{\text{stag}} \in \mathbb{R}^{d_{stag}}$.
We obtain probability distribution over the POS tags and supertags by:
\begin{align}
   \text{softmax}(W^{\text{(pos)}}h_k^{\text{pos}}+b^{\text{(pos)}})\label{pos:eq}
   \\
   \text{softmax}(W^{\text{(stag)}}h_k^{\text{stag}}+b^{\text{(stag)}})
   \label{stag:eq}
\end{align}
where $W^{\text{(pos)}}$, $b^{\text{(pos)}}$,  $W^{\text{(stag)}}$, and $b^{\text{(stag)}}$ are in $\mathbb{R}^{n_{pos} \times d_{pos}}$, $\mathbb{R}^{n_{pos}}$, $\mathbb{R}^{n_{stag} \times d_{stag}}$, and $\mathbb{R}^{n_{stag}}$ respectively, with  $n_{pos}$ and $n_{stag}$ the numbers of possible POS tags and supertags respectively.

We use the same hyperparameters as in the parser.  The MLPs for POS tagging and supertagging both contain 500 units.
We again train this model with the Adam algorithm to minimize the sum of the cross-entropy losses from head predictions ($s_i$ from Eq.\ \ref{unlabeled:eq}), label predictions ($\ell_i$ from Eq.\ \ref{labeled:eq}), POS predictions (Eq.\ \ref{pos:eq}), and supertag predictions (Eq.\ \ref{stag:eq}) with $\ell=0.01$ and batch size 100. 
After each training epoch, we test the parser on the dev set and compute the percentage of each token that is assigned the correct parent, relation, supertag, and POS tag.
When the percentage does not improve on five consecutive epochs, training ends.

This joint modeling has several advantages. 
First, the joint model yields a full syntactic analysis simultaneously without the need for training separate models or performing jackknife training.
Secondly, joint modeling introduces a bias on the hidden representations that could allow for better generalization in each task  \cite{caruana1997}.
Indeed,  in experiments described in a later section, we  show empirically that predicting POS tags and supertags does indeed benefit performance on parsing (as well as the tagging tasks).  

\section{Results and Discussion}
We follow the protocol of \citet{bangalore-EtAl:2009:NAACLHLT09-Short}, \citet{chung:etal:TAG+12:2016}, \citet{kasai-EtAl:2017:EMNLP2017}, and \citet{friedman-EtAl:2017:TAG+13}; we use the grammar and the TAG-annotated WSJ  Penn Tree Bank  extracted by  \citet{chen05}. 
Following that work, we use Sections 01-22 as the training set,  Section 00 as the dev set, and Section 23 as the test set. 
The training, dev, and test sets comprise 39832, 1921, and 2415 sentences, respectively. We implement all of our models in TensorFlow \cite{abadi2016tensorflow}.\footnote{Our code is available online for easy replication of our results at \url{https://github.com/jungokasai/graph_parser}.}
\subsection{Supertaggers}
Our BiLSTM POS tagger yielded 97.37\% and 97.53\% tagging accuracy on the dev and test sets, performance on par with the state-of-the-art \cite{wang:2015, ma-hovy:2016:P16-1}.\footnote{We cannot directly compare these results because the data split is different in the POS tagging literature.} Seen in the middle section of Table \ref{stagging-results} is supertagging performance obtained from various model configurations. 
``Final concat'' in the model name indicates that vectors from forward and backward pass are concatenated only after the final layer. Concatenation happens after each layer otherwise. Numbers immediately after BiLSTM indicate the numbers of layers. CNN, HW, and POS denote respectively character-level CNNs, highway connections, and pipeline POS input from our BiLSTM POS tagger. 
Firstly, the  differences in performance between BiLSTM2 (final concat) and BiLSTM2 and between BiLSTM2 and BiLSTM2-CNN suggest an advantage to performing concatenation after each layer and adding character-level CNNs. Adding predicted POS to the input somewhat helps supertagging though the difference is small. Adding a third BiLSTM layer helps only if there are highway connections, presumably because deeper BiLSTMs are more vulnerable to the vanishing/exploding gradient problem.
Our  supertagging model (BiLSTM3-HW-CNN-POS) that performs best on the dev set achieves an accuracy of 90.81\% on the test set, outperforming the previously best result by more than 1.3\%.
\begin{table}
\small
\centering
\begin{tabular}{ l |cc }
\hline
  Supertagger & Dev & Test \\\hline\hline
  \citet{bangalore-EtAl:2009:NAACLHLT09-Short} &88.52& 86.85\\
  \citet{chung:etal:TAG+12:2016} &87.88& --\\
  \citet{kasai-EtAl:2017:EMNLP2017} &89.32& 89.44\\\hline
  BiLSTM2 (final concat)&88.96& --\\
  BiLSTM2&89.60& --\\
  BiLSTM2-CNN&89.97 & --\\
  BiLSTM2-CNN-POS& 90.03&-- \\
  BiLSTM2-HW-CNN-POS& 90.12&-- \\
  BiLSTM3-CNN-POS& 90.12& --\\
  BiLSTM3-HW-CNN-POS& \underbar{90.45}& \underbar{90.81}\\
  BiLSTM4-CNN-POS&89.99 &--\\
  BiLSTM4-HW-CNN-POS& 90.43& --\\\hline
  
  Joint (Stag) & 90.51&--\\
  Joint (POS+Stag) & \textbf{90.67}&\textbf{91.01}\\\hline
\end{tabular}
\caption{Supertagging Results. Joint (Stag) and Joint (POS+Stag) indicate joint parsing models that perform supertagging, and POS tagging and supertagging respectively.}
\label{stagging-results}
\end{table}
%

\begin{table}
\small
\centering
\begin{tabular}{ l |cc }
\hline
  POS tagger & Dev & Test \\\hline\hline
  BiLSTM &97.37&97.53\\
  Joint (POS+Stag) &97.54 &97.73
\end{tabular}
\caption{POS tagging results.}
\label{pos-results}
\end{table}

\begin{table}[t]
\hspace{-4mm}
\small
\centering
\begin{tabular}{ l |cc|cc}
\hline
    & \multicolumn{2}{c|}{Dev} &\multicolumn{2}{c}{Test}\\
  Parser &UAS&LAS&UAS&LAS\\\hline\hline
  \citet{bangalore-EtAl:2009:NAACLHLT09-Short} &87.60&85.80&86.66&84.90\\
  \citet{chung:etal:TAG+12:2016} &89.96&87.86&--&--\\
  \citet{friedman-EtAl:2017:TAG+13} &90.36&88.91&90.31&88.96\\
  \citet{kasai-EtAl:2017:EMNLP2017} &90.88&89.39&90.97&89.68\\\hline
  BiLSTM3&91.75&90.22&--&--\\ 
  BiLSTM3-CNN&92.27&90.76&--&--\\ 
  BiLSTM3-CNN-POS&92.07&90.53&--&--\\  
  BiLSTM3-CNN-Stag&92.15&90.65&--&--\\  
  BiLSTM3-HW-CNN&92.29&90.71&--&--\\ 
  BiLSTM4-CNN&92.11&90.66&--&--\\ 
  BiLSTM4-HW-CNN&\underbar{92.78}&\underbar{91.26}&92.77&91.37\\ 
  BiLSTM5-CNN&92.34&90.77&--&--\\ 
  BiLSTM5-HW-CNN&92.64&91.11&--&--\\ 
  \hline
  Joint (Stag)&92.97&91.48&--&--\\ 
  Joint (POS+Stag)&\textbf{93.22}&\textbf{91.80}&\textbf{93.26} &\textbf{91.89}\\ 
  \hline
  Joint (Shuffled Stag) &92.23&90.56&--&--\\\hline
\end{tabular}
\caption{Parsing results on the dev and test sets.}
\label{parsing-results}
\end{table}

\subsection{Parsers}
Table \ref{parsing-results} shows parsing results on the dev set. Abbreviations for models are as before with one addition:
Stag denotes pipeline supertag input from our best supertagger (BiLSTM3-HW-CNN-POS in Table \ref{stagging-results}).
As with supertagging, we observe a gain from adding character-level CNNs. 
Interestingly, adding predicted POS tags or supertags deteriorates performance with BiLSTM3.
These results suggest that morphological information and word information from character-level CNNs and word embeddings overwhelm the information from predicted POS tags and supertags.
Again, highway connections become crucial as the number of layers increases. 
We finally evaluate the parsing model with the best dev performance (BiLSTM4-HW-CNN) on the test set (Table \ref{parsing-results}). 
It achieves 91.37 LAS points and 92.77 UAS points, improvements of 1.8 and 1.7 points respectively from the state-of-the-art.

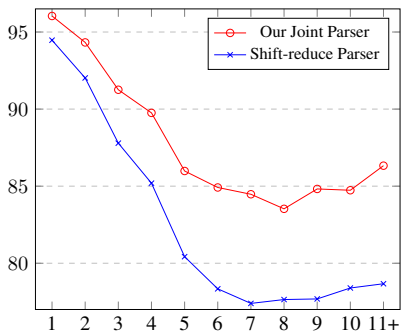
\begin{figure}[b]
\centering
\begin{tikzpicture}[scale=0.7]
\begin{axis}[
    title={},
    xmin=0, xmax=220,
    ymin=77.0, ymax=96.5,
    xtick={10,30,50,70,90,110, 130, 150, 170, 190, 210},
    xticklabels={1, 2, 3, 4, 5, 6, 7, 8, 9, 10, 11+},
    ytick={80, 85, 90, 95},
    legend pos=north east,
    legend style={font=\footnotesize},
    ymajorgrids=true,
    grid style=dashed,
]
 
\addplot[
    color=red,
    mark=o,
    ]
    coordinates {(10,96.04)(30,94.325)(50,91.255)(70,89.755)(90,85.985)(110,84.915)(130,84.475)(1
50,83.53)(170,84.82)(190,84.735)(210,86.33)};
    \addlegendentry{Our Joint Parser}
\addplot[
    color=blue,
    mark=x,
    ]
    coordinates {(10,94.47)(30,92.02)(50,87.79)(70,85.195)(90,80.425)(110,78.34)(130,77.395)(150,
77.645)(170,77.685)(190,78.395)(210,78.67)};
    \addlegendentry{Shift-reduce Parser}
\end{axis}
\end{tikzpicture}
\caption{F1 Score with Dependency Length.}
\label{f1_dep_length}
\end{figure}

\begin{figure}[b]
\hspace{-4mm}
\centering
\begin{tikzpicture}[scale=0.7]
\begin{axis}[
    title={},
    xmin=0, xmax=220,
    ymin=87.0, ymax=98.0,
    xtick={10,30,50,70,90,110, 130, 150, 170, 190, 210},
    xticklabels={1, 2, 3, 4, 5, 6, 7, 8, 9, 10, 11+},
    ytick={85, 90, 95},
    legend pos=north east,
    legend style={font=\footnotesize},
    ymajorgrids=true,
    grid style=dashed,
]
 
\addplot[
    color=red,
    mark=o,
    ]
    coordinates {(10,96.62)(30,93.44)(50,93.08)(70,91.91)(90,91.59)(110,92.565)(130,92.865)(150,94.01)(170,92.09)(190,90.725)(210,91.78)};
    \addlegendentry{Our Joint Parser}
\addplot[
    color=blue,
    mark=x,
    ]
    coordinates {(10,92.995)(30,90.075)(50,90.31)(70,89.605)(90,89.01)(110,88.935)(130,89.63)(150,90.35)(170,89.875)(190,90.005)(210,88.11)};
    \addlegendentry{Shift-reduce Parser}
\end{axis}
\end{tikzpicture}
\caption{F1 Score with Distance to Root.}
\label{f1_distance_to_root}
\end{figure}
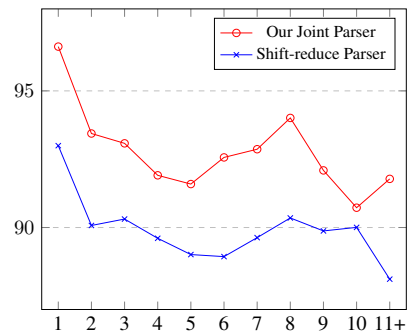

\subsection{Joint Models}
We provide joint modeling results for supertagging and parsing in Tables  \ref{pos-results} and \ref{parsing-results}. For these joint models, we employed the best parsing configuration (4 layers of BiLSTMs, character-level CNNs, and highway connections), with and without POS tagging added as an additional task.
We can observe that our full joint model that performs POS tagging, supertagging, and parsing further improves performance in all of the three tasks, yielding the test result of 91.89 LAS and 93.26 UAS points, an improvement of more than 2.2 points each from the state-of-the-art.


%

Figures \ref{f1_dep_length} and \ref{f1_distance_to_root} illustrate the relative performance of the feed-forward neural network shift-reduce TAG parser \cite{kasai-EtAl:2017:EMNLP2017} and our joint graph-based parser with respect to two of the measures explored by \citet{McDonald:2011}, namely dependency length and distance between a dependency and the root of a parse.
The graph-based parser  outperforms the shift-reduce parser across all conditions. Most interesting is the fact that the graph-based parser shows less of an effect of dependency length. Since the shift-reduce parser builds a parse sequentially with one parsing action depending on those that come before it, we would expect to find a propogation of errors made in establishing shorter dependencies to the establishment of longer dependencies.

Lastly, it is worth noting our joint parsing architecture has a substantial advantage regarding parsing speed.
Since POS tagging, supertagging, and parsing decisions are made independently for each word in a sentence, our system can  parallelize computation once the sentence is encoded in the BiLSTM layers.
Our current implementation processes 225 sentences per second on a single Tesla K80 GPU, an order of magnitude faster than the MICA system \cite{bangalore-EtAl:2009:NAACLHLT09-Short}.\footnote{While such computational resources were not available in 2009, our parser differs from the MICA chart parser in being able to better exploit parallel computation enabled by modern GPUs.}

\section{Joint Modeling and Network Representations}

Given the improvements we have derived from the joint models, we analyze the nature of inductive bias that results from  multi-task training and attempt to provide  an explanation as to why joint modeling improves performance.

\subsection{Noise vs.\ Inductive Bias}

One might argue that joint modeling improves performance merely because it adds noise to each task and prevents over-fitting.
If the introduction of noise were the key, we would still expect to gain an improvement in parsing even if the target supertag were corrupted, say by shuffling the order of supertags for the entire training data \cite{caruana1997}. 
We performed this experiment, and the result is shown as ``Joint (Shuffled Stag)" in Table~\ref{parsing-results}. Parsing performance falls behind the best non-joint parser by 0.7 LAS points.
This suggests that inducing the parser to create  representations to predict both supertags and a parse tree is beneficial for both tasks, beyond a mere introduction of noise.

\subsection{Syntactic Analogies}

We next analyze the induced vector representations in the output projection matrices of our supertagger and joint parsers using the syntactic analogy framework \cite{kasai-EtAl:2017:EMNLP2017}.
Consider, for instance, the analogy that an elementary tree representing a clause headed by a transitive verb (t27) is to a clause headed by an intransitive verb (t81) as a subject relative clause headed by a transitive verb (t99) is to a subject relative headed by an intransitive verb (t109). 
Following  the ideas in \citet{Mikolov2013} for word analogies, we can express this structural analogy as {t27 - t81 + t109 = t99} and test it by cosine similarity.
Table \ref{analogy-results} shows the results of the analogy test with 246  equations involving structural analogies with only the 300 most frequent supertags in the training data.
While the embeddings (projection matrix) from the independently trained supertagger do not appear to reflect the syntax, those obtained from the joint models yield linguistic structure despite the fact that the supertag embeddings (projection matrix) is trained without any {\em a priori} syntactic knowledge about the elementary trees. 

The best performance is obtained by the supertag representations obtained from the training of the transition-based parser \citet{kasai-EtAl:2017:EMNLP2017} and \citet{friedman-EtAl:2017:TAG+13}.  For the transition-based parser, it is beneficial to share statistics among the input supertags that differ only by a certain operation or property \cite{kasai-EtAl:2017:EMNLP2017} during the training phase, yielding the success in the analogy task. 
For example, a transitive verb supertag whose object has been filled by substitution should be treated by the parser in the same way as an intransitive verb supertag.
In our graph-based parsing setting, we do not have a notion of parse history or partial derivations that directly connect intransitive and transitive verbs.  
However, syntactic analogies still hold to a considerable degree in the vector representations of supertags induced by our joint models, with average rank of the correct answer nearly the same as that obtained in the transition-based parser.

%
This analysis bolsters our hypothesis that joint training biases representation learning toward linguistically sensible structure.
The supertagger is just trained to predict linear sequences of supertags. 
In this setting, many intervening supertags can occur, for instance, between a subject noun and its verb, and the supertagger might not be able to systematically link the presence of the two in the sequence.
In the joint models, on the other hand, parsing actions will explicitly guide the network to associate the two supertags.

\begin{table}
\small
\centering
\begin{tabular}{ l |ccc}
\hline
   Parser / Supertagger&\%correct&Avg.\ rank\\\hline\hline
   Transition-based&67.07&2.36\\
   Our Supertagger &0.00&152.46\\
   Our Joint (Stag)&29.27&\underbar{2.55}\\
   Our Joint (POS+Stag)&\underbar{30.08}&2.57\\
\end{tabular}
\caption{Syntactic analogy test results on the 300 most frequent supertags. Avg.\ rank is the average position of the correct choice in the ranked list of the closest neighbors; the top line indicates the result of using supertag embeddings that are trained jointly with a transition based parser
\cite{friedman-EtAl:2017:TAG+13}.}
\label{analogy-results}
\end{table}


\section{Downstream Tasks}
Previous work has applied TAG parsing to the downstream tasks of syntactically-oriented textual entailment \cite{xu-EtAl:2017:TAG+13} and semantic role labeling \cite{chenrambow2003}. 
In this work, we apply our parsers to the textual entailment and unbounded dependency recovery tasks and achieve state-of-the-art performance.
These results bolster the significance of the improvements gained from our joint parser and the utility of TAG parsing for downstream tasks.
\subsection{PETE}

Parser Evaluation using Textual Entailments
(PETE) is a shared task from the SemEval-2010
Exercises on Semantic Evaluation \cite{yuret-han-turgut:2010:SemEval}. 
The task was intended to evaluate syntactic
parsers across different formalisms, focusing
on entailments that could be determined entirely
on the basis of the syntactic representations
of the sentences that are involved, without recourse to lexical semantics, logical reasoning, or world knowledge. 
For example, syntactic knowledge
alone tells us that the sentence \textit{John, who
loves Mary, saw a squirrel} entails \textit{John saw a squirrel} and \textit{John loves Mary} but not, for instance, that \textit{John knows Mary} or \textit{John saw an animal}. 
Prior work found the best performance was achieved with parsers using grammatical frameworks that provided rich linguistic descriptions, including CCG \cite{cambridge, schwa}, Minimal Recursion Semantics (MRS) \cite{mrs-system}, and TAG \cite{xu-EtAl:2017:TAG+13}.
\citet{xu-EtAl:2017:TAG+13} provided a set of linguistically-motivated transformations to use TAG derivation trees to solve the PETE task. We follow their procedures and evaluation for our new parsers. 

We present test results from two configurations in Table~\ref{pete-results}. One configuration is a pipeline approach that runs our BiLSTM POS tagger, supertagger, and parser.
The other one is a joint approach that only uses our full joint parser.
The joint method yields 78.1\% in accuracy and 76.4\% in F1, improvements of 2.4 and 2.7 points over the previously reported best results.
\begin{table}
\small
\centering
\begin{tabular}{ l |cccc}
\hline
   System &\%A&\%P&\%R&F1\\\hline\hline
   \citet{cambridge}&72.4&79.6& 62.8& 70.2\\
   \citet{schwa}&70.4 &68.3&\bf{80.1}&\underbar{73.7}\\
   \citet{mrs-system}&70.7&\bf{88.6}& 50.0&63.9\\
   \citet{xu-EtAl:2017:TAG+13}&\underbar{75.7}&88.1&61.5 &72.5\\\hline
   Our Pipeline Method&77.1&86.6&66.0&74.9\\
   Our Joint Method&\textbf{78.1} &86.3&68.6 &\textbf{76.4} \\ 
\end{tabular}
\caption{PETE test results. Precision (P), recall (R), and F1 are calculated for ``entails."}
\label{pete-results}
\end{table}

\begin{table*}[t]
\vspace{-4mm}
\small
\centering
\begin{tabular}{ l |ccccccccc}
\hline
   System&ObRC&ObRed&SbRC&Free&ObQ&RNR&SbEm&Total&Avg\\\hline\hline
   C\&C (CCG)& 59.3&62.6&80.0&72.6&\bf{72.6}&\bf{49.4}&22.4&53.6&\underbar{61.1}\\
   Enju (HPSG)&\underbar{47.3}&\underbar{65.9}&82.1&\underbar{76.2}&32.5&47.1&32.9&\underbar{54.4}&54.9\\
   Stanford (PCFG) &22.0&1.1&74.7&64.3&41.2&45.4&10.6&38.1&37.0\\
   MST (Stanford Dependencies)&34.1&47.3&78.9&65.5&41.2&45.4&\underbar{37.6}&49.7&50.0 \\
   MALT (Stanford Dependencies)&40.7&50.5&\textbf{84.2}&70.2&31.2&39.7&23.5&48.0&48.5\\\hline
   NN Shift-Reduce TAG Parser &60.4&75.8&68.4&79.8&53.8&45.4&44.7&59.4&61.2\\
   Our Joint Method&  \bf{72.5}&\bf{78.0}&81.1&\bf{85.7}&\underbar{56.3}&\underbar{47.1}&\bf{49.4}&\bf{64.9}&\bf{67.0}\\
\end{tabular}
\caption{Parser accuracy on the unbounded dependency corpus. The results of the first five parsers are taken from \citet{unbounded-dependencies} and \citet{nivre-EtAl:2010:PAPERS}. The Total and Avg columns indicate the percentage of correctly recovered dependencies out of all dependencies and the average of accuracy on the 7 constructions.}
\label{unbounded-results}
\end{table*}

\subsection{Unbounded Dependency Recovery}
The unbounded dependency corpus \cite{unbounded-dependencies} specifically evaluates parsers on unbounded dependencies, which involve a constituent moved from its original position, where an unlimited number of clause boundaries can intervene.
The corpus comprises 7 constructions:
object extraction from a relative clause (ObRC),
object extraction from a reduced relative clause
(ObRed), subject extraction from a relative clause
(SbRC), free relatives (Free), object wh-questions (ObQ), right node raising (RNR), and subject extraction
from an embedded clause (SbEm).

Because of variations across formalisms in their representational format for unbounded depdendencies, past work has conducted manual evaluation on this corpus \cite{unbounded-dependencies, nivre-EtAl:2010:PAPERS}.
 We instead conduct an automatic evaluation using a procedure that converts TAG parses to structures directly comparable to those specified in the unbounded dependency corpus.
To this end, we apply  two types of structural transformation in addition to those used for the PETE task:\footnote{One might argue that since the unbounded dependency evaluation is recall-based, we added too many edges by the transformations. However, it turns out that applying all the transformations for the corpus even improves performance on PETE (77.6 F1 score), which considers precision and recall, verifying that our transformations are reasonable.} 1) a more extensive analysis of coordination, 2) resolution of differences in dependency representations in cases involving copula verbs and co-anchors (e.g., verbal particles). See Appendix \ref{sec:supplemental} for  details. After the transformations, we simply check if the resulting dependency graphs contain target labeled arcs given in the dataset.

Table \ref{unbounded-results} shows the results. Our joint parser outperforms the other parsers, including the neural network shift-reduce TAG parser \cite{kasai-EtAl:2017:EMNLP2017}.
Our data-driven parsers yield relatively low performance in the ObQ and RNR constructions. Performance on ObQ is low, we expect, because of their rarity in the data on which the parser is trained.\footnote{The substantially better performance of the C\&C parser is in fact the result of additions that were made to the training data.} For RNR, rarity may be an issue as well as the limits of the TAG analysis of this construction.  Nonetheless, we see that the rich structural representations that a TAG parser provides enables substantial improvements in the extraction of unbounded dependencies. In the future, we hope to evaluate state-of-the-art Stanford dependency parsers automatically.
\section{Related Work}

The two major classes of data-driven methods for dependency parsing are often called transition-based and graph-based parsing \cite{dependencyparsing2009}.
Transition-based parsers (e.g. MALT \cite{Nivre2003AnEA}) learn to predict the next transition given the input and the parse history.
Graph-based parsers (e.g. MST \cite{mcdonald-EtAl:2005:HLTEMNLP}) are trained to directly assign scores to dependency graphs.

Empirical studies have shown that a transition-based parser and a graph-based parser yield similar overall performance across languages \cite{McDonald:2011}, but the two strands of data-driven parsing methods manifest the fundamental trade-off of parsing algorithms.
The former prefers rich feature representations with parsing history over global training and exhaustive search, and the latter allows for global training and inference at the expense of limited feature representations \cite{dependencyparsing2009}.

Recent neural network models for transition-based and graph-based parsing can be viewed as remedies for the aforementioned limitations.
\citet{andor-EtAl:2016:P16-1} developed a transition-based parser using feed-forward neural networks that performs global training approximated by beam search.
The globally normalized objective addresses the label bias problem and makes global training effective in the transition-based parsing setting.
\citet{KG2016} incorporated a dynamic oracle \cite{GoldbergNivre2013} in a BiLSTM transition-based parser that remedies global error propagation. 
\citet{KG2016} and \citet{dozatmanning2017} proposed graph-based parsers that have access to rich feature representations obtained from BiLSTMs.

Previous work integrated CCG supertagging and parsing using belief propagation and dual decomposition approaches \cite{P11-1048}.
\citet{K17-3014} incorporated a graph-based dependency parser \cite{KG2016} with POS tagging.
Our work followed these lines of effort and improved TAG parsing performance.


%

\section{Conclusion and Future Work}
In this work, we presented a state-of-the-art TAG supertagger, a parser, and a joint parser that performs POS tagging, supertagging, and parsing.
The joint parser has the benefit of giving a full syntactic analysis of a sentence simultaneously.
Furthermore, the joint parser achieved the best performance, an improvement of over 2.2 LAS points from the previous state-of-the-art.
We have also seen that the joint parser yields state-of-the-art in textual entailment and unbounded dependency recovery tasks, and raised the possibility that TAG can provide useful structural analysis of sentences for other NLP tasks.
We will explore more applications of our TAG parsers in future work.

\bibliography{naaclhlt2018}
\bibliographystyle{acl_natbib}

\appendix
\section{Transformations for Unbounded Dependency Recovery Corpus}
\label{sec:supplemental}

For automatic evaluation on the unbounded dependency recovery corpus (UDR, \citet{unbounded-dependencies}), we run simple conversion of dependency labels in UDR to those in our TAG grammar (See Table \ref{conversion}) with a couple of exceptions. 
\begin{itemize}
    \item Change arcs from verbs to wh-adverbs as in ``where is the city located?" to adjunction.
    \item Reflect causative-inchoative alternation in the subject embedded construction. Concretely, change the role of ``door" in ``hold the door shut" from the subject to the object of ``shut.''
\end{itemize}
We then transform TAG dependency trees. 
Finally, we simply check if the resulting dependency graphs contain target labeled arcs given in the dataset.

\begin{table}[b]
\vspace{-4mm}
\small
\centering
\begin{tabular}{ l |c}
\hline
  UDR Labels & TAG Labels\\\hline
  nsubj, cop& 0\\
  dobj, pobj, obj2, nsubjpass& 1\\ 
  others (advmod etc) & ADJ\\ 
  \end{tabular}
\caption{UD to TAG label conversion. }
\label{conversion}
\end{table}
Below is a full description of transformations.
This set of structural transformations is applied
in the order in which we will present it, so that
the output of previous transformations can feed
subsequent ones.
In the following, we denote an arc pointing from node B to node A with label C as (A, B, C) where A and B are called the child (dependent) and the parent (head) in the relation.
\subsection{Transformations from PETE}
We apply three types of transformation from \citet{xu-EtAl:2017:TAG+13} to interpret the TAG parses.
\paragraph{Relative Clauses}
When an elementary tree of a relative clause  adjoins into a noun, we add a reverse arc with the label reflecting the type of the relative clause elementary tree.
For a subject relative, we
add a 0-labeled arc, for an object relative, we add a 1-labeled arc, and so forth.
\paragraph{Sentential Complements}
Sentential complementation in TAG derivations
can be analyzed via either adjoining the higher
clause into the embedded clause (necessarily so in
cases of long-distance extraction from the embedded
clause) or substituting the embedded clause in
the higher clause.
In order to normalize this divergence, for an adjunction arc involving a predicative auxiliary elementary tree (supertag), we add a reverse arc involving the 1 relation (sentential complements).
\subsection{Coordination}
We roughly follow the method presented in \citet{xu-EtAl:2017:TAG+13} with extensions.
Under the TAG analysis, VP coordination
involves a VP-recursive auxiliary tree headed
by the coordinator that includes a VP substitution
node (for the second conjunct) with label 1.
In order to allow the first clause’s subject
argument (as well as modal verbs and negations)
to be shared by the second verb, we add the
relevant relations to the second verb.
In addition, we analyze sentential coordination cases. 
Sentence coordination in our TAG grammar usually happens between two complete sentences and no modifiers or arguments are shared, and therefore it can be analyzed via substituting a sentence int the coordinator with label 1.
However, when sentential coordination happens between two relative clause modifiers, our TAG grammar analyzes the second clause as a complete sentence, meaning that we need to recover the extracted argument by consulting the property of the first clause.
Furthermore, the deep syntactic role of the extracted argument can be different in the two relative clauses. 
For instance, in the sentence, ``... the same stump which had impaled the car of many a guest in the past thirty years and which he refused to have removed," we need to recover an arc from removed to stump with label 1 whereas the arc from impaled to stump has label 0. 
To resolve this issue, when there is coordination of two relative clause modifiers, we add an edge from the head of the second clause to the modified noun with the same label as the label that under which the relative pronoun is attached to the head. 
\subsection{Resolving Differences in Dependency Representations}
\paragraph{Small Clauses}
The UDR corpus has inconsistency with regards to small clauses.
UDR gives an analysis that a small clause contains a subject and a complement as in (nsubj, guy, liar) in ``the guy who I call a liar." in the subject embedded constructions.
However, in the object question and object free relative constructions, a small clause is analyzed as two arguments of the verb. For instance, UDR specifies (what, adopted, dobj) in ``we adopted what I would term pseudo-capitalism." 
To solve this problem we add an arc from the head of the matrix clause to the subject in a small clause with label 1. 

\begin{figure}
\hspace{-4mm}
    \centering
    \includegraphics[width=0.45\textwidth]{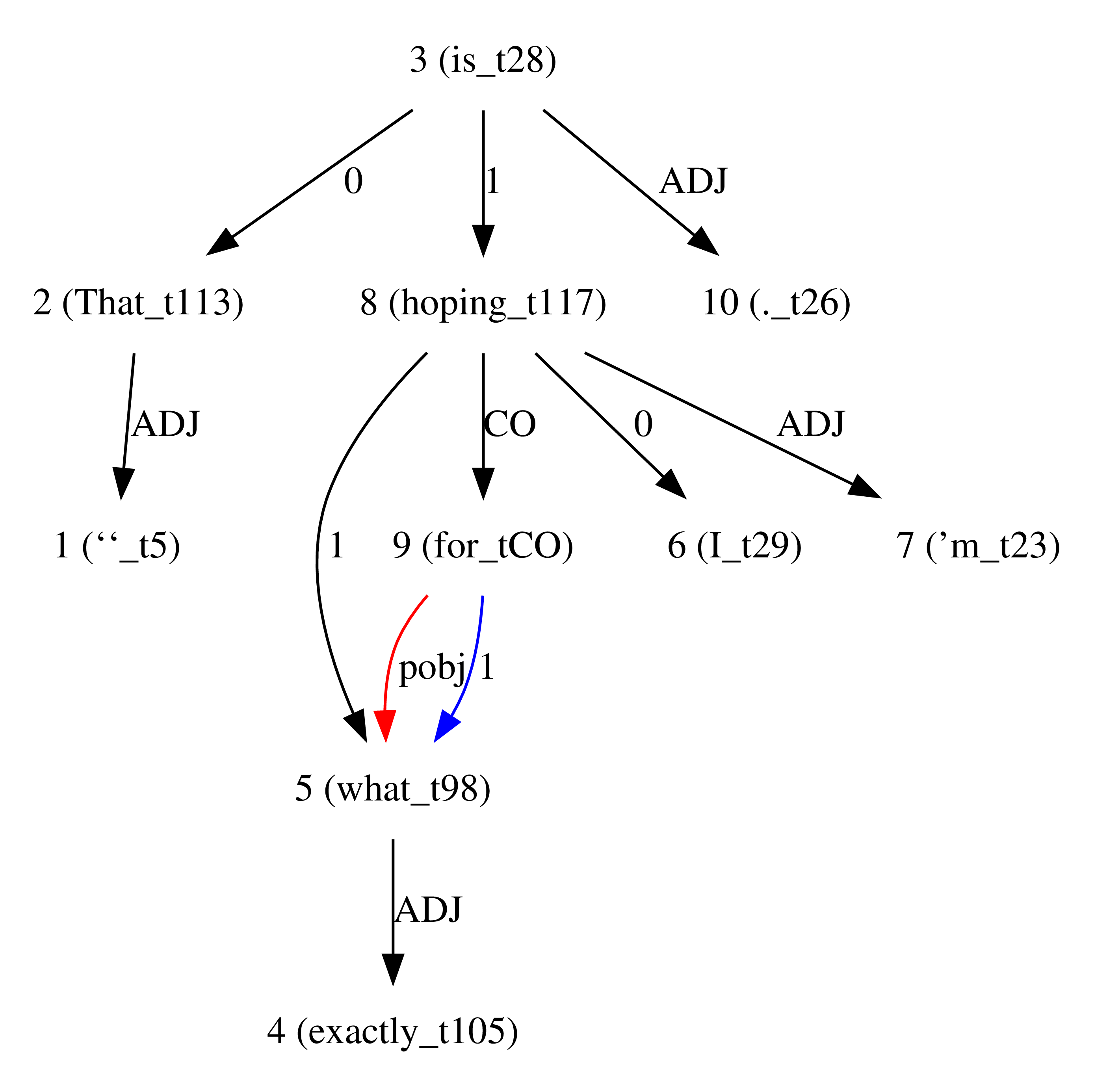}
    \caption{Co-anchor case from a sentence \textit{``that is exactly what I'm hoping for.} The UDR gives the red arc (what, for, pobj). The blue arc (what, for, 1) is obtained from (what, hope, 1).}
    \label{coanchor}
\end{figure}

\begin{figure}
\hspace{-4mm}
    \centering
    \includegraphics[width=0.45\textwidth]{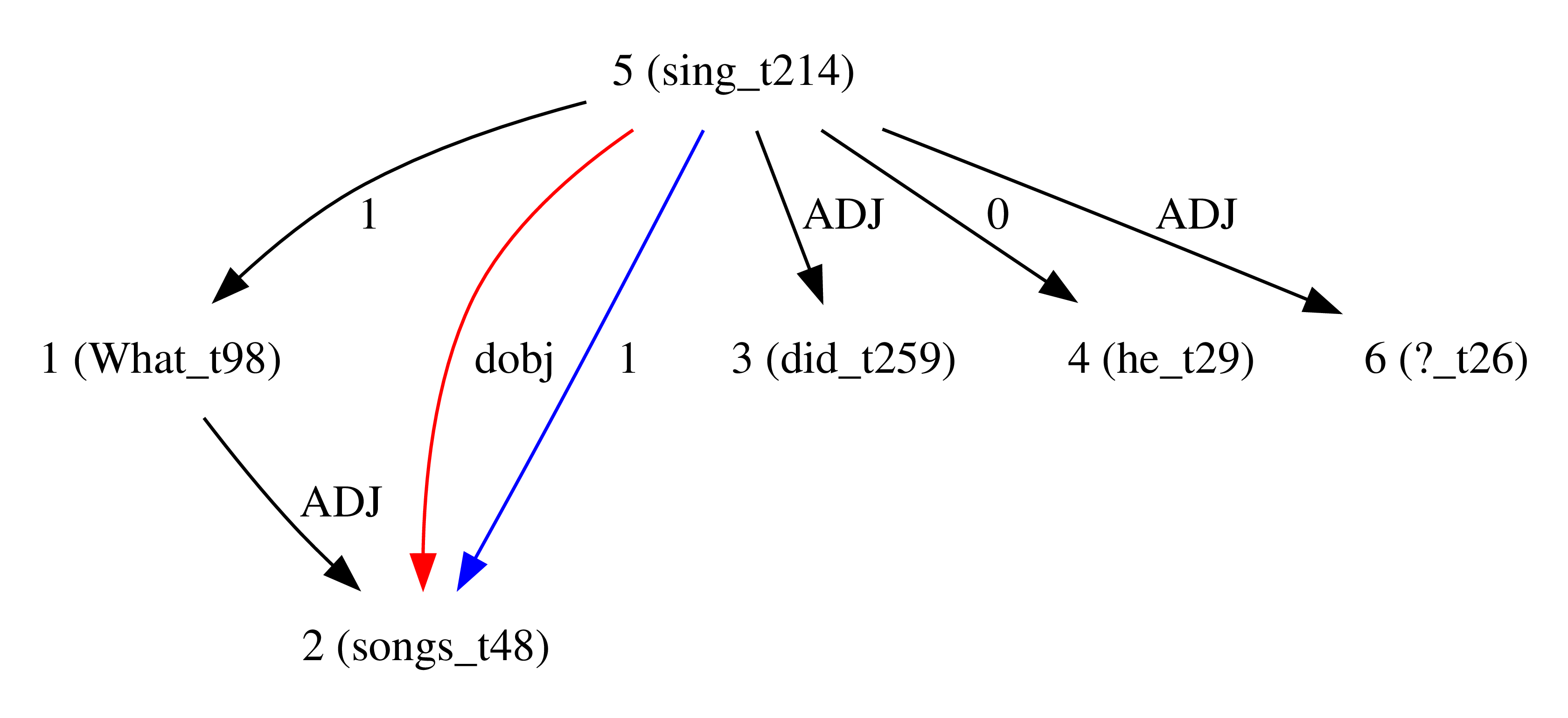}
    \caption{Wh-determiner case from a sentence \textit{What songs did he sing?} The UDR gives the red arc (songs, sing, dobj). The blue arc (song, sing, 1) is obtained from (what, sing, 1) and (songs, what, ADJ).}
    \label{wh_adj}
\end{figure}

\paragraph{Co-anchors}
In our TAG grammar, Co-anchor attachment represents the
substitution into a node that is construed as a co-head of an elementary tree. For instance, ``for" is deemed as a co-anchor to ``hope" in the sentence \textit{``that is exactly what I'm hoping for} (Figure \ref{coanchor}).
In this case,  UDR would pick the relation (what, hope, pobj).
Therefore, when there is a co-anchor to a head tree, we add all arcs that involve the head tree to the co-anchor tree.
\paragraph{Wh-determiners and Wh-adverbs}
Our TAG grammar analyzes a wh-determiner via adjoining the noun into the wh-determiner (Figure \ref{wh_adj}).
This is also true for cases where a wh-adverb is followed by an adjective and a noun as in \textit{\textbf{how many battles} did she win?}
In contrast, UDR corpus gives an analysis that the noun is the head of the constituent.
In order to resolve this discrepancy, when a word adjoins into a wh-word,\footnote{We considered imposing a more strict condition that the word adjoining into the wh-word is a noun, but we found cases that this method fails to cover; for example, UDR gives (dobj, get, much) for a sentence ``opinion is mixed on how \textbf{much} of a boost the overall stock market would \textbf{get} even if dividend growth continues at double-digit levels."} we pick all arcs with the wh-word as the child and add the arcs obtained from such arcs by replacing the wh-word child by the word adjoining into the wh-word.


\paragraph{Copulas}
A copula is usually treated as a dependent to the predicate both in our TAG grammar (adjunction) and UDR. However, we found two situations where they differ from each other. First, when wh-extraction happens on the complement, as in ``obviously there has been no agreement on what American conservatism is, or rather, what it should be," the TAG grammar analyzes it via substituting the wh-word (``what") into the copula (``is").  
To reconcile this disagreement between the TAG grammar and UDR, when substitution happens into a {\em be} verb, we add the substitution into the copula.\footnote{We use the nltk lemmatizer \cite{nltk} to identify {\em be} verbs.} 
Second, UDR treats non-{\em be} copulas differently than {\em be} verbs.
An example is the UDR relation (those, stayed, nsubj) ``in the other hemisphere it is growing colder and nymphs, those who stayed alive through the summer, are being brought into nests for quickening and more growing'' where our parser yields (those, alive, 0).
For this reason, when a lemma of a verb is a non-\textit{be} copula,\footnote{We chose ``
stay," ``become," ``seem," and ``remain."} we add arcs involving the word to the copula adjoining into the copula.
\paragraph{PP attachment with multiple noun candidates}
We observed that PP attachment with multiple noun candidates is often at stake in UDR.\footnote{This is indeed one of the problems with UDR. Performance on UDR is not purely reflective of unbounded dependency recovery.} 
For instance,  
UDR provides (part, had, nsubj) and (several, tried, nsubj) for the sentences ``... there is no part of the earth that has not had them" and ``there were several on the Council who tried to live like Christians" while the TAG parser outputs (earth, had, nsubj) and (Council, tried, nsubj) respectively.
While we count these cases as ``wrong" since they manifest certain disambiguation (though not purely unbounded dependency recovery), we ignore superficial (conventional) differences in head selection.
In our TAG grammar ``a lot of people" would be headed by ``lot" whereas UDR would recognize ``people" as the head.  
Hence, when ``lot/lots/kind/kinds/none of" occurs, we add all arcs with ``lot/lots/kind/kinds/none" to the head of the phrase that is the object of ``of."

\paragraph{Modals} 
In the UDR corpus, a modal depends on an auxiliary verb following the modal, if there is one.
For example, ``Rosie reinvented this man, who may or may not have known about his child" is given the relation (may, have, aux).
In the TAG grammar, both ``may" and ``have" adjoin into ``known."
Therefore, when the head of a modal has another child with adjunction, we add an arc from the child to the modal.
\paragraph{Existential {\em there}}
UDR gives the ``cop" relation between an existential there and the \textit{be} verb.
For example, it gives (be, legislation, cop) in ``... on how much social legislation there should be.''
On the other hand, our TAG grammar analyzes that ``there" is attached to ``be" with label 0.\footnote{Usually, ``there" is attached to the noun, not the \textit{be} verb, but in this case, extraction is happening on the noun, so the \textit{be} verb becomes the head. See the discussion on copulas above.}
To resolve this issue, for arcs that point into an existential there with label 0, we add a reverse edge with label 0. 

\paragraph{Determiner modifying a sentence}
Finally, when a determiner followed by an adjective modifies a sentence via adjunction in our TAG as in ``the more highly placed they are -- that is, the more they know -- the more concerned they have become," we add an edge from the verb to the adjective with label 1.
\end{document}